\documentclass[conference,a4paper]{IEEEtran}
\IEEEoverridecommandlockouts
\usepackage[T1]{fontenc}
\usepackage{cite}
\usepackage{amsmath,amssymb,amsfonts}
\usepackage{graphicx}
\usepackage{textcomp}
\usepackage{xcolor}
\usepackage{booktabs}
\usepackage{multirow}
\usepackage{tikz}
\usepackage{url}
\usetikzlibrary{positioning,calc,arrows.meta}

\def\BibTeX{{\rm B\kern-.05em{\sc i\kern-.025em b}\kern-.08em
    T\kern-.1667em\lower.7ex\hbox{E}\kern-.125emX}}


\begin{document}

\title{Different Corruptions, Different Signals:\\
Uncertainty and Loss in Federated Data Quality}

\author{
\IEEEauthorblockN{Bradley Scott, Zeqi Luo, and Edmond S. L. Ho}
\IEEEauthorblockA{\textit{School of Computing Science, University of Glasgow}\\
Glasgow, United Kingdom\\
Bradley.Scott@glasgow.ac.uk, z.luo.3@research.gla.ac.uk, Shu-Lim.Ho@glasgow.ac.uk}
}

\maketitle

\begin{abstract}
Federated learning (FL) data corruption can affect either inputs or labels, but it
remains unclear whether input-conditional uncertainty and prediction-label loss
expose these corruption modes equally. This paper compares two corruption-detection signals in FL: input-conditional uncertainty and prediction-label loss. The uncertainty signal is characterised using a learned aleatoric variance estimate together with Monte Carlo (MC) dropout variance and entropy measures, while the loss is computed against the supplied label. We test these signals against additive image noise
and persistent random label flips. On ResNet-20 with CIFAR-10 and SVHN under Dirichlet partitions with data that are not independent and identically distributed (non-IID), the two corruption
types behave differently. For persistent random label flips, the within-client per-sample area under the receiver operating characteristic curve (AUC) is 0.85 on CIFAR-10 and 0.95 on SVHN for prediction-label loss, while every uncertainty estimator stays at chance (0.49--0.50). This pattern is consistent with the model remaining confident in the underlying image despite the supplied label being wrong. For image noise,
expected-entropy uncertainty rises above chance (0.67 on CIFAR-10 and 0.66 on SVHN), while loss responds comparably (0.64 on both). Each signal is therefore the stronger detector
for a different corruption: the prediction-label loss for persistent label flips, and
expected-entropy uncertainty for image noise, with its advantage becoming apparent as federation-wide corruption prevalence increases. Robust FL data-quality assessment should match the signal to the corruption rather than rely on uncertainty alone across corruption types.
\end{abstract}


\section{Introduction}

Federated learning (FL) trains a shared model across clients without centralising their
raw data~\cite{mcmahan2017}. That privacy boundary also makes data-quality assessment
harder: the server sees updates and model outputs, not the examples that produced them.
Recent robust FL methods use uncertainty to identify noisy clients or inspect suspicious
updates. FedNed~\cite{fedned2024} and FedDPSO~\cite{feddpso2025} use predictive uncertainty to identify extremely
noisy clients before changing how those clients affect training.
SureFED addresses poisoning-robust FL differently, using uncertainty-aware model
evaluation to inspect client updates~\cite{surefed2023}. Together, these methods make
uncertainty a natural candidate for detecting corrupted data in FL.

FL data corruption is not a single problem. Image noise changes the input $x$: the label
is unchanged, but the image is degraded. A persistent label flip changes the supplied
label $y$: the image is unchanged, but the same wrong label is used whenever that sample
is seen. These corruptions affect different parts of the training pair $(x,y)$ and should
not be expected to leave the same evidence in the model outputs.

This matters because two different corruption-detection signals are often discussed together. Input-conditional uncertainty scores, such as a learned aleatoric variance estimate~\cite{kendall2017}, expected entropy~\cite{depeweg2018}, or Monte Carlo (MC) dropout disagreement~\cite{gal2016}, ask whether the model is unsure about the image. Loss scores computed against the supplied label ask whether the model believes the label attached to that image. This is the kind of prediction-label signal used by label-error methods such as
Confident Learning~\cite{confident2021}. For input corruption these two questions can point in the same direction. This motivates a direct test: if corruption changes the image, uncertainty and loss may both respond; if corruption changes only the label, loss may respond even when image-prediction uncertainty does not.

This gives our research question: \emph{In FL, do uncertainty and loss detect different
forms of data corruption?} We compare persistent random label flips
with additive image noise, and we report both per-client and within-client per-sample
detection area under the receiver operating characteristic curve (AUC). This tests whether uncertainty and loss are interchangeable corruption
detectors, or whether each detects a different corruption mode. The per-client view
matches the way many FL systems rank clients, but it is confounded by between-client
heterogeneity, including class-composition differences in client data that are not
independent and identically distributed (non-IID). The within-client test is the decisive one: corrupted and clean samples are
compared inside the same client and under the same global model, so partition effects are
held fixed.

We evaluate ResNet-20 on CIFAR-10 and SVHN under 10-client Dirichlet non-IID FL, sweeping
both the corruption rate and the number of corrupt clients. The results support a
multi-signal view. For persistent label flips, the input-conditional uncertainty scores stay at chance in the averaged within-client per-sample results (AUC
$0.49$--$0.50$), while the prediction-label loss separates flipped samples ($0.85$ on
CIFAR-10 and $0.95$ on SVHN). For image noise, expected-entropy uncertainty rises above chance
(per-sample AUC $0.67$ on CIFAR-10 and $0.66$ on SVHN), while the loss responds comparably ($0.64$ on both). The practical conclusion is
not that one signal replaces the other, but that each signal is informative for a
different part of the corruption space.

The paper makes three contributions. First, it compares uncertainty and loss as
corruption-detection signals for two data-corruption modes in FL: image noise and
persistent label flips. Second, it shows why per-client detection is not enough by adding
a within-client per-sample test that controls for between-client heterogeneity. Third, it shows that signal informativeness depends on both corruption type and
federated prevalence: persistent label corruption is better detected by the
prediction-label loss, while the expected-entropy advantage for image noise becomes apparent as federation-wide corruption prevalence increases.
\section{Related Work}

\subsection{Predictive uncertainty and prediction-label signals}
Predictive uncertainty is commonly discussed in terms of aleatoric (data) and epistemic
(model) components~\cite{kendall2017}. In this paper we instantiate that distinction using variance and entropy measures. The variance pair comprises a learned aleatoric estimate and MC dropout probability variance, while the entropy pair comprises expected entropy and mutual information from the MC dropout predictive distributions~\cite{gal2016,depeweg2018}.
The uncertainty scores we evaluate are \emph{input-conditional}: they are computed from the input through the model without using the supplied label at probe time. Separately, noisy-label methods often use
the model probability assigned to the \emph{given} label to identify likely annotation errors;
Confident Learning~\cite{confident2021} is the canonical example of this prediction-label mismatch signal. Notably, the noisy-label detectors that work in
practice rank by such prediction-label signals (e.g.\ UNICON's Jensen--Shannon divergence
to the label~\cite{unicon2022}) or by epistemic/predictive uncertainty together with
Confident Learning~\cite{uqled2024}. Where a learned aleatoric estimate is included, its role is to stabilise training rather than to select samples~\cite{ulc2022}. We evaluate both the
input-conditional uncertainty signals and the prediction-label mismatch signal, and keep
them distinct.

\subsection{Label noise and corruption in FL}
Robust noisy-label methods in FL are predominantly sample-selection or
label-correction schemes; the FNBench benchmark surveys this family across many
methods and noise patterns but contains no aleatoric/epistemic uncertainty
treatment~\cite{fnbench2025}, leaving open what uncertainty can
detect about FL label noise. Dedicated label-flipping defences abandon uncertainty
entirely and detect label-flipping attacks from output-layer gradients~\cite{lfighter2024}. FedBary studies federated data valuation and reports a
detection \emph{asymmetry}, in which feature corruption is easy to detect and
label corruption is hard, but via optimal-transport (Wasserstein) distances rather than
uncertainty~\cite{fedbary2024}. We instead test this asymmetry directly by comparing input-conditional uncertainty
with prediction-label loss under both corruption types.

\subsection{Established uses of uncertainty in FL}
Most FL uses of uncertainty target ends where its value is well established: Bayesian
personalisation and calibration~\cite{fedivon}, predictive-uncertainty model
selection~\cite{fedpn2024}, selective prediction and misclassification
detection~\cite{fedee}, fairness client weighting~\cite{udjfl2025}, and uncertainty-aware
inspection of poisoned updates~\cite{surefed2023}. These are \emph{input}-side and
\emph{model}-side judgements (out-of-distribution detection, abstention, routing, calibration), not
label-quality assessment.

\subsection{Uncertainty for noisy-client detection}
FedNed identifies \emph{extremely} noisy clients ($>90\%$ label noise) using MC dropout
uncertainty~\cite{fedned2024}, while FedDPSO similarly targets ${\sim}90\%$-noise clients
before optimising aggregation weights~\cite{feddpso2025}. Both evaluate client models on
server-side auxiliary data within mixed-noise federations. Our corrupt clients instead
have $20$--$60\%$ persistent random flips, and we evaluate a common global model on local
probes. We therefore ask which of the two signals, input-conditional uncertainty or prediction-label loss, reveals moderate within-client corruption.

\section{Method}

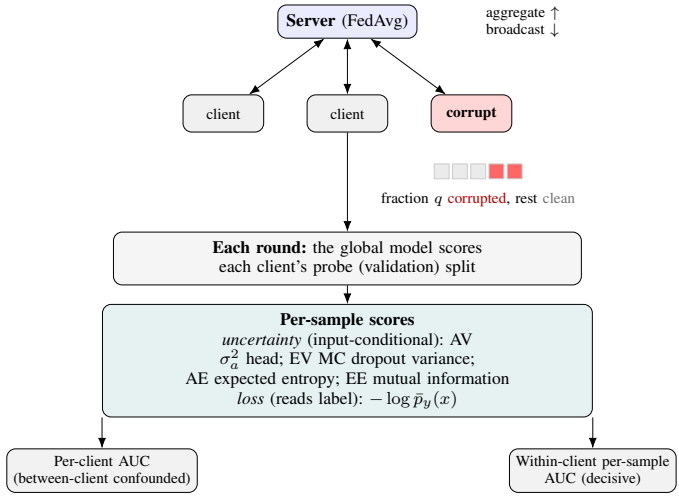
\begin{figure}[t]\centering
\resizebox{\columnwidth}{!}{%
\begin{tikzpicture}[font=\footnotesize,>=Latex,
  srv/.style={draw,rounded corners,fill=blue!8,align=center,inner sep=4pt},
  cli/.style={draw,rounded corners,align=center,inner sep=2pt,minimum width=13mm,minimum height=6mm,font=\scriptsize},
  proc/.style={draw,rounded corners,align=center,inner sep=4pt},
  res/.style={draw,rounded corners,fill=black!5,align=center,inner sep=3pt,font=\scriptsize},
  sq/.style={draw=black!20,minimum size=2.4mm,inner sep=0pt}]
 \node[srv] (srv) {\textbf{Server} (FedAvg)};
 \node[cli,fill=black!6] (c1) [below=9mm of srv,xshift=-20mm] {client};
 \node[cli,fill=black!6] (c2) [below=9mm of srv] {client};
 \node[cli,fill=red!16]  (c3) [below=9mm of srv,xshift=20mm] {\textbf{corrupt}};
 \draw[<->] (srv)--(c1); \draw[<->] (srv)--(c2); \draw[<->] (srv)--(c3);
 \node[align=left,font=\scriptsize] at ([xshift=17mm]srv.east) {aggregate $\uparrow$\\broadcast $\downarrow$};
 \node[sq,fill=black!8] (m1) [below=5mm of c3,xshift=-4.8mm] {};
 \node[sq,fill=black!8,right=0.4mm of m1] (m2) {};
 \node[sq,fill=black!8,right=0.4mm of m2] (m3) {};
 \node[sq,fill=red!60,right=0.4mm of m3] (m4) {};
 \node[sq,fill=red!60,right=0.4mm of m4] (m5) {};
 \node[font=\scriptsize,below=1mm of m3] {fraction $q$ \textcolor{red!70!black}{corrupted}, rest \textcolor{black!55}{clean}};
 \node[proc,fill=gray!8,text width=0.82\columnwidth] (probe) [below=16mm of c2] {\textbf{Each round:} the global model scores each client's probe (validation) split};
 \draw[->] (c2)--(probe);
\node[proc,fill=teal!10,text width=0.86\columnwidth,align=center] (sig) [below=3mm of probe]
  {\textbf{Per-sample scores}\\
   \emph{uncertainty} (input-conditional): AV $\sigma_a^2$ head; EV MC dropout variance;\\
   AE expected entropy; EE mutual information\\
   \emph{loss} (reads label):  $-\log \bar p_y(x)$};
\draw[->] (probe)--(sig);
 \node[res] (pc) [below=5mm of sig.south west,anchor=north] {Per-client AUC\\(between-client confounded)};
 \node[res] (ps) [below=5mm of sig.south east,anchor=north] {Within-client per-sample\\AUC (decisive)};
 \draw[->] (sig.south west)--(pc); \draw[->] (sig.south east)--(ps);
\end{tikzpicture}}
\caption{Measurement pipeline. A subset of the $10$ clients is corrupt (count $M$ swept):
a fraction $q$ of a corrupt client's samples are corrupted (flips \emph{or} noise; separate conditions), the
rest clean. The global model probes every client each round; uncertainty and loss are recorded as diagnostic signals and scored per-client and within-client per-sample.}
\label{fig:pipeline}
\end{figure}

Fig.~\ref{fig:pipeline} sketches the measurement pipeline. We
use federated training with Federated Averaging (FedAvg) aggregation over $N$ clients,
of which $M$ are corrupt. Within each corrupt client, samples are corrupted at rate $q$ by persistent
label flips \emph{or} additive image noise (studied as separate conditions),
and the remaining samples stay clean, with the assignment fixed by sample
index every round. Each round, the
current \emph{global} model probes every client's local data and records, per sample,
four input-conditional uncertainty signals (AV from a learned aleatoric head, and EV, AE, and EE from MC dropout predictive distributions) alongside the prediction-label loss. Uncertainty is never used to weight or filter clients; the evaluated scores are used as diagnostic signals rather than server-side selection or aggregation rules. We then score how well each signal separates corrupt from clean
data at two granularities: across clients (per-client, which between-client
heterogeneity can confound) and within a single corrupt client (which controls for
between-client partition and model differences). The
following subsections define the signals, the estimators, the corruption mask, and these
detection metrics in turn.

\subsection{Signals}
Each client evaluates the current global model on its local validation split (a
\emph{probe}) and records, per sample, the prediction-label loss together with
four input-conditional uncertainty signals. We
represent the aleatoric/epistemic distinction in two forms: a variance-based
pair (AV/EV) and an entropy-based pair (AE/EE). AV is obtained from a learned,
dropout-free aleatoric head, whereas EV, AE, and EE are statistics computed
from repeated MC dropout predictive distributions.

\subsubsection{Variance signals}
The aleatoric variance (AV) is the direct output of the learned aleatoric head,
$\sigma_a^2(x)$. This head contains no dropout. The epistemic variance (EV),
by contrast, is computed from $T$ stochastic softmax predictions obtained with
MC dropout: for each input, we calculate the per-class probability variance
across the $T$ predictions and average it over classes~\cite{gal2016}. Thus AV
is a learned per-sample variance estimate, whereas EV measures variation across
repeated stochastic predictions. This variance-based pair follows the aleatoric/epistemic construction of UDF-GMA~\cite{udfgma2025}. During
training, AV and EV are combined by a learned fusion function for loss
attenuation; at probe time, however, AV and EV are recorded separately as
diagnostic signals.

\subsubsection{Entropy signals}
The $T$ MC dropout predictive distributions also provide two entropy-based
uncertainty signals, neither of which uses the learned aleatoric variance head.
Aleatoric entropy (AE) is the expected entropy of the stochastic predictions,
capturing uncertainty within each prediction and averaging it across passes.
Epistemic entropy (EE) is the mutual-information term, capturing disagreement
between stochastic predictions~\cite{depeweg2018}. EE is exactly the
\emph{Bayesian Active Learning by Disagreement} (BALD) acquisition
score~\cite{bald2011}; here it is used purely as an epistemic-uncertainty
measure and no active learning is performed.

\subsubsection{Prediction-label loss}
The loss signal is the only evaluated per-sample score that reads the supplied
label. It is $\ell=-\log \bar p_y(x)$, where $\bar p_y(x)$ denotes the
probability assigned to the supplied label $y$ by the mean of the $T$
MC dropout predictive distributions. Thus $e^{-\ell}=\bar p_y(x)$, so ranking samples by high loss is
equivalent to ranking them by low self-confidence, as used by Confident
Learning~\cite{confident2021}.

\subsection{Estimators and the loss-attenuation mechanism}
Let $\phi(x)$ be the backbone features, $\mu_t(x)$ the logits from the
dropout prediction head on the $t$-th dropout pass, and
$p_t(x)=\mathrm{softmax}(\mu_t(x))$. The learned aleatoric head predicts a
per-sample non-negative uncertainty scalar
\begin{equation}
\sigma_a^2(x)=\mathrm{softplus}\!\big(f_a(\phi(x))\big)\in\mathbb{R}_{>0},
\label{eq:sigma}
\end{equation}
where $f_a$ is a small learned, dropout-free multilayer perceptron head. At
inference, $\sigma_a^2(x)$ depends on the image, not the supplied label,
although it is learned through the label-dependent objective below.

In the reported runs the attenuation loss combines this aleatoric estimate
with the MC dropout probability variance,
\begin{align}
e(x)&=\frac{1}{C}\sum_c\mathrm{Var}_t[p_{t,c}(x)],\nonumber\\
\tau^2(x)&=f_\sigma\!\big(\sigma_a^2(x),e(x)\big),
\label{eq:tau}
\end{align}
where $f_\sigma$ is a learned softplus fusion head and $C$ is the number of
classes. Thus $\sigma_a^2(x)$ is the aleatoric-variance signal (AV) and $e(x)$
the epistemic-variance signal (EV), while $\tau^2(x)$ is the scalar variance
used for attenuation. At probe time, AV is obtained directly from the
dropout-free aleatoric head, whereas EV, AE, and EE are computed from $T$
MC dropout predictive distributions.

During training, we use $K$ MC dropout passes to form the mean logits
$\bar\mu(x)$ and independently use $K$ Gaussian draws to approximate the
attenuated predictive probability $\tilde p(x)$, with
$\tau(x)=\sqrt{\tau^2(x)}$; the same sample count is used for both:
\begin{align}
\bar\mu(x)&=\tfrac1K\sum_{i=1}^K\mu_i(x),\nonumber\\
\tilde p(x)&=\tfrac1K\!\sum_{j=1}^{K}
\mathrm{softmax}\!\big(\bar\mu(x)+\tau(x)\,\varepsilon_j\big),
\;\; \varepsilon_j\!\sim\!\mathcal N(0,I),\label{eq:attn}\\
\mathcal L_{\text{a}}(x,y)&=-\log \tilde p_y(x)
+\lambda\big(e^{\tau^2(x)}-1\big),\label{eq:la}
\end{align}
where $\mu_i(x)$ and $p_i(x)=\mathrm{softmax}(\mu_i(x))$ denote the
logits and predictive distribution from the $i$-th of the $K$
training-time MC dropout passes, while $j$ indexes the independent
Gaussian draws $\varepsilon_j$ used for loss attenuation. The overall
objective is
\begin{equation}
\mathcal L=
\frac{1}{K}\sum_{i=1}^{K}\big[-\log p_{i,y}(x)\big]
+\beta\,\mathcal L_{\text{a}},
\label{eq:objective}
\end{equation}
a plain per-pass cross-entropy plus the attenuated term
$\mathcal L_{\text{a}}$, where $\beta$ balances the attenuated term against the plain loss and
$\lambda$ weights the variance penalty within it. These correspond to
UDF-GMA's $\lambda_1$ and $\lambda_0$, respectively, and both are set
to $1$ as in that work. The penalty discourages large $\tau^2$, while the likelihood term can
favour variance that reduces prediction-label loss; this trade-off does not
guarantee AV calibration.

The entropy-based diagnostic signals are
\begin{equation}
\underbrace{
\frac{1}{T}\sum_{t=1}^{T}\mathcal H[p_t(x)]
}_{\text{aleatoric (AE)}},
\qquad
\underbrace{
\mathcal H\!\left[\frac{1}{T}\sum_{t=1}^{T}p_t(x)\right]
-\frac{1}{T}\sum_{t=1}^{T}\mathcal H[p_t(x)]
}_{\text{epistemic (EE)}},
\label{eq:bald}
\end{equation}
where $\mathcal H[\cdot]$ denotes Shannon entropy. AE measures uncertainty
within individual stochastic predictions, averaged across passes, whereas EE
measures disagreement between the stochastic predictions. Neither signal reads
the supplied label.

\subsection{Why flips should evade input-conditional uncertainty}
The signal definitions motivate, but do not determine, an asymmetry because corrupted
labels also affect optimisation. Each flipped sample receives a fixed, randomly chosen incorrect label. Because these wrong labels are distributed across the other classes rather than concentrated on one alternative class, the underlying class can remain the dominant label signal even at substantial client-level corruption. Input-conditional uncertainty can therefore fail to rank flipped samples while prediction-label loss remains high. Class-consistent flips would instead concentrate corruption on one wrong label and could change this relationship. Sufficiently long training may also fit individual corrupted labels~\cite{arpit2017}. Image noise instead perturbs the input itself, so
input-conditional estimators can respond directly.

\subsection{Per-condition corruption mask}
In each condition, samples in a corrupt client are corrupted at rate $q$ using
a deterministic index-seeded mask; the remaining samples stay clean. The corruption type
is set by the condition: a label flip (the sample is given a randomly chosen incorrect label) in the \emph{flip} condition, and additive Gaussian image noise in the \emph{noise} condition. Flip and noise are studied as separate conditions, so a corrupt client is never both flipped and noised. The selected sample indices are fixed across rounds and compared runs; flip labels are persistent, while Gaussian perturbations are resampled. This gives per-sample ground truth for sample-level detection: each corrupt
client holds clean and corrupted samples scored by the same global model, so the per-sample
test below separates them.

\subsection{Detection metrics}
For each signal we report a Mann--Whitney detection AUC: the probability that a randomly
chosen \emph{corrupt} item is scored as more suspicious than a randomly chosen \emph{clean}
one,
\begin{equation}
\mathrm{AUC}=\frac{1}{|P|\,|Q|}\sum_{i\in P}\sum_{j\in Q}
\Big[\mathbf{1}(s_i>s_j)+\tfrac12\mathbf{1}(s_i=s_j)\Big],
\label{eq:auc}
\end{equation}
where $P$ and $Q$ are the corrupt and clean item sets, respectively, and $s_i$
is the score assigned to item $i$. It counts only \emph{cross}
(corrupt-vs-clean) pairs -- order \emph{within} either group is irrelevant,
and no item is compared against another in its own group -- so $0.5$ indicates
chance-level ranking and values below $0.5$ indicate inverse ranking.
We compute it at two granularities that differ in what $P$ and $Q$ are, and
crucially in what they hold fixed.

\subsubsection{Per-client} Each client's score is its signal averaged over its
samples; $P$ contains the $M$ corrupt clients and $Q$ the $N-M$ clean clients. This ranks clients
\emph{across} the federation, so it absorbs \emph{any} systematic difference between the two groups,
including non-IID class composition and other between-client differences. A corruption
signal here is therefore \emph{confounded}: a client can be ranked apart because of
pre-existing client heterogeneity as well as because the corruption is visible. As a
clean-data control, we also compute the same AUC on corresponding clean runs, using as
$P$ the client IDs designated as corrupt in the matched corruption runs; all clients
in this control are clean, so any separation reflects pre-existing between-client
differences rather than corruption.

\subsubsection{Per-sample} Within a \emph{single} corrupt client, $P$ is that client's corrupted samples
and $Q$ its clean samples, scored by the \emph{same} global model. Because $P$ and $Q$ come from the same client and are scored by the same global
model, client identity, the client-level partition, and model state are held fixed. The
within-client AUC therefore controls for the between-client confounding present in
the per-client comparison, although residual differences in class composition or
intrinsic sample difficulty between the corrupted and clean subsets may remain. Within each round, the within-client AUC is computed separately for each corrupt client and then averaged equally across clients. The per-client AUC is reported to expose that between-client confounding.

\subsection{Measure-only protocol}
Aggregation is fixed to FedAvg; uncertainty affects local optimisation through the
objective above but is never used to weight or gate clients. Probe scores and
within-client AUCs are computed locally; the server receives only scalar client
metrics, not raw samples, labels, or per-sample scores.

\section{Experiments}

\subsection{Setup}
\begin{figure}[!t]
\centerline{\includegraphics[width=\columnwidth]{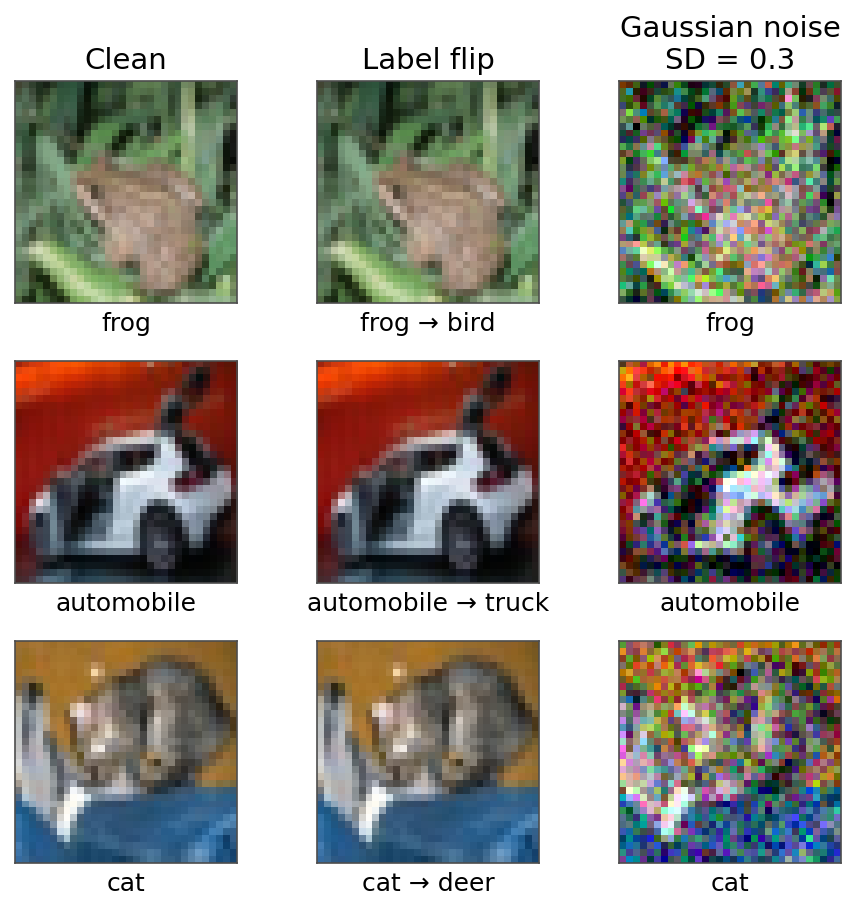}}
\caption{CIFAR-10 examples: a clean sample, a persistent label flip (unchanged image, incorrect label), and additive Gaussian image noise (SD $0.3$). Flip and noise are applied as separate conditions.}
\label{fig:corruption-main4}
\end{figure}

We use ResNet-20~\cite{he2016} with uncertainty components based on
UDF-GMA~\cite{udfgma2025}, evaluated on CIFAR-10~\cite{krizhevsky2009}
and SVHN~\cite{netzer2011}. Both datasets use $3\times32\times32$ RGB
inputs and otherwise follow the same experimental protocol. We use $N=10$ clients, full participation, 50 rounds, FedAvg aggregation, and the local objective combining cross-entropy with attenuation from Section~III-B. We use Dirichlet
non-IID skew $\alpha\in\{0.1,0.3,1.0\}$ and $T=20$ MC dropout probe passes.
Conditions are \emph{clean}, persistent random label \emph{flip}, and additive
Gaussian image \emph{noise} (standard deviation (SD) $0.3$; Fig.~\ref{fig:corruption-main4}); flip and noise are run as separate conditions. Corrupt-client sets are reused across
conditions so signals are directly comparable. We sweep the per-client corruption
rate $q\in\{0.2,0.4,0.6\}$ and corrupt-client count $M\in\{2,4,6,8\}$ of $10$,
giving a $q\times M$ grid. Each of the $25$ conditions is run at all three
$\alpha$ values and three random seeds, where the seed fixes the client-partition
draw, model initialisation, and corruption assignment, giving $9$ runs per condition:
$225$ per dataset and $450$ in total. Reported metrics use rounds $41$--$50$: mean AUC and maximum accuracy. For seed-level variability, AUCs are averaged
equally over the $M\times q\times\alpha$ grid within each seed, with sample SD
across the three seed-specific averages. The attenuation loss uses $K=3$ for both the MC dropout passes and Gaussian
MC samples. All federated simulations use Flower~1.29~\cite{beutel2020flower}. Code and experiment scripts are publicly available on GitHub.\footnote{\url{https://github.com/bs97/different-corruptions-different-signals}}

\subsection{Per-client detection is confounded by client heterogeneity}
Ranking clients by a signal (are corrupt clients placed above clean ones?) does not
isolate corruption from pre-existing between-client differences. Table~\ref{tab:perclient} shows the resulting mixed picture (full grid across $M$ and $q$ in Table~\ref{tab:perclient-full}, Appendix~A): the
aleatoric variance signal is below chance on flips, while the entropy signals can rank corrupt clients above clean ones, especially on SVHN. The label-dependent error signal ranks flip clients cleanly, but that ranking is still a between-client comparison, so it can reflect client heterogeneity as well as corruption. Stratifying the per-client AUC by Dirichlet $\alpha$ shows that these rankings
vary substantially with the partitioning regime
(Table~\ref{tab:perclient-alpha}, Appendix~A). Despite this variation,
$1{-}\mathrm{Acc}$ remains strong for flips across all $\alpha$ values
($0.91$--$1.00$), so the client-level prediction-label separation is not explained
solely by between-client heterogeneity. A clean-data control nevertheless confirms that
client-level ranking can reflect pre-existing between-client differences: using
the same client subsets designated as corrupt in the matched corruption runs,
but their scores from corresponding clean runs with no corruption present,
produced AUCs ranging from $0.36$ to $0.67$ on CIFAR-10 and from $0.45$ to
$0.66$ on SVHN. Accuracy deficit and loss are both prediction-label signals: each compares the model's prediction with the supplied label, which the uncertainty scores never do. We report the accuracy deficit at the client level and the loss at the sample level. Image noise is more visible at the client level, but that result is still mixed with between-client effects. The within-client test below controls for these between-client differences.

\begin{table}[!t]
\caption{Per-client corruption-detection AUC, averaged over the full $M\times q\times\alpha\times$seed grid. All columns are detection AUCs. AV/EV: aleatoric/epistemic variance; AE/EE: aleatoric/epistemic entropy; $1{-}\mathrm{Acc}$ scores clients by inverse global-probe accuracy.}
\label{tab:perclient}
\begin{center}
\small
\setlength{\tabcolsep}{2.5pt}
\begin{tabular*}{\columnwidth}{@{\extracolsep{\fill}}ll ccccc}
\toprule
Dataset & Corrupt & AV & AE & EV & EE & $1{-}$Acc \\
\midrule
CIFAR-10 & Flip  & 0.41 & 0.63 & 0.55 & 0.58 & \textbf{0.96} \\
CIFAR-10 & Noise & 0.45 & \textbf{0.85} & 0.70 & 0.75 & 0.73 \\
SVHN     & Flip  & 0.45 & 0.73 & 0.61 & 0.63 & \textbf{0.99} \\
SVHN     & Noise & 0.22 & \textbf{0.90} & 0.82 & 0.83 & \textbf{0.90} \\
\bottomrule
\end{tabular*}
\end{center}
\end{table}

\subsection{Within-client per-sample detection}
Within a single corrupt client we ask, per sample, whether each signal separates the
flipped samples or image-noise samples from the clean ones -- a test that holds client-level partition and model state fixed
(Table~\ref{tab:persample}). The result is consistent
across both datasets. In the grid-averaged results, every
input-conditional signal -- variance and entropy, aleatoric and epistemic -- sits at chance on
flipped samples ($0.49$--$0.50$). The prediction-label loss separates them
($0.85$ on CIFAR-10 and $0.95$ on SVHN), with the full grid
(Table~\ref{tab:persample-full}, Appendix~A) showing the same pattern across $M$ and $q$. On image noise, AE achieved a modestly higher grid-averaged AUC than loss ($0.67$/$0.66$ versus $0.64$). Across the three seed-specific grid averages, the paired AE--loss difference was
$0.029\pm0.012$ on CIFAR-10 and $0.014\pm0.002$ on SVHN
(mean $\pm$ SD; Table~\ref{tab:noise-seed-values}, Appendix~A). The full grid (Table~\ref{tab:persample-full}, Appendix~A) shows that once
$M\ge4$, AE exceeds loss in every noise cell on both datasets (by up to
$0.07$ on CIFAR-10). Because $q$ is the within-client corruption rate, increasing $M$ also increases
the overall amount of corrupted data in the federation.
The AE advantage therefore becomes apparent as federation-wide corruption
prevalence increases. AV is weak on CIFAR-10 noise ($0.52$) and systematically inverted on SVHN noise
($0.28$); the cause is unresolved, so entropy carries the positively oriented uncertainty result.

\begin{table}[!t]
\caption{Within-client per-sample corruption-detection AUC, averaged over the full $M\times q\times\alpha\times$seed grid. AV/EV: aleatoric/epistemic variance; AE/EE: aleatoric/epistemic entropy; $L$ is prediction-label loss.}
\label{tab:persample}
\begin{center}
\small
\setlength{\tabcolsep}{3pt}
\begin{tabular*}{\columnwidth}{@{\extracolsep{\fill}}ll ccccc}
\toprule
Dataset & Corrupt & AV & AE & EV & EE & L \\
\midrule
CIFAR-10 & Flip  & 0.49 & 0.49 & 0.50 & 0.50 & \textbf{0.85} \\
CIFAR-10 & Noise & 0.52 & \textbf{0.67} & 0.60 & 0.61 & 0.64 \\
SVHN     & Flip  & 0.49 & 0.50 & 0.50 & 0.50 & \textbf{0.95} \\
SVHN     & Noise & 0.28 & \textbf{0.66} & 0.62 & 0.62 & 0.64 \\
\bottomrule
\end{tabular*}
\end{center}
\end{table}

\subsection{The corruption missed by uncertainty is more harmful}
Persistent label flips sharply reduce clean-test performance. Across the $M\times q$ grid
(Table~\ref{tab:harm}), flip accuracy falls as $M$ and $q$ increase, reaching $31.1\%$
on CIFAR-10 ($39$ points below the $69.7\%$ clean baseline) and $59.5\%$ on SVHN
($34$ points below $93.2\%$) at $M{=}8,q{=}0.6$. Yet per-sample uncertainty on flips
remains at chance ($0.49$--$0.50$). Image noise, which expected entropy detects,
changes accuracy by only $+0.2$ to $+1.4$ points, consistent with augmentation.
Thus expected-entropy uncertainty is more responsive to the lower-impact input
corruption than to the harmful label corruption.

\subsection{The loss erodes as corruption prevalence increases}
The prediction-label loss detects flips, but its AUC falls as corruption prevalence increases, from $0.93$/$0.98$ at the mildest setting to $0.70$/$0.86$ at
$M{=}8,q{=}0.6$ (CIFAR-10/SVHN). This is consistent with training assigning increasing probability to corrupted labels, which reduces the prediction-label mismatch that loss exploits. Because our random flips are dispersed across alternative classes rather than concentrated on one wrong label, however, the underlying class can remain the dominant label signal. Loss therefore weakens but remains above chance throughout the tested grid.

\begin{table}[!t]
\caption{Best clean-test accuracy (\%) over rounds $41$--$50$, averaged over
$\alpha$ and seeds. $M$: corrupt clients (of $10$); $q$: corrupted fraction within a corrupt client.}
\label{tab:harm}
\begin{center}
\small
\setlength{\tabcolsep}{2pt}
\begin{tabular*}{\columnwidth}{@{\extracolsep{\fill}}ll ccc ccc}
\toprule
 & & \multicolumn{3}{c}{CIFAR-10} & \multicolumn{3}{c}{SVHN} \\
\cmidrule(lr){3-5}\cmidrule(lr){6-8}
Corruption & $M$ & $q{=}0.2$ & $q{=}0.4$ & $q{=}0.6$ & $q{=}0.2$ & $q{=}0.4$ & $q{=}0.6$ \\
\midrule
Clean & $0$ & \multicolumn{3}{c}{$69.7$} & \multicolumn{3}{c}{$93.2$} \\
\midrule
\multirow{4}{*}{Flip}
 & $2$ & 66.8 & 65.9 & 64.6 & 92.4 & 92.0 & 91.6 \\
 & $4$ & 63.1 & 58.5 & 55.2 & 90.1 & 88.6 & 86.8 \\
 & $6$ & 57.4 & 50.3 & 43.4 & 86.2 & 81.1 & 75.4 \\
 & $8$ & 52.1 & 40.7 & 31.1 & 82.4 & 73.1 & 59.5 \\
\midrule
\multirow{4}{*}{Noise}
 & $2$ & 70.0 & 70.0 & 70.4 & 93.4 & 93.5 & 93.4 \\
 & $4$ & 70.2 & 70.8 & 70.8 & 93.4 & 93.5 & 93.6 \\
 & $6$ & 70.7 & 71.1 & 70.8 & 93.4 & 93.5 & 93.7 \\
 & $8$ & 70.7 & 70.7 & 70.8 & 93.6 & 93.7 & 93.6 \\
\bottomrule
\end{tabular*}
\end{center}
\end{table}

\section{Discussion and Limitations}

We evaluate ResNet-20, 10 clients, FedAvg, two datasets (CIFAR-10 and SVHN),
persistent random flips, and one Gaussian-noise level; corruption is data-level (mislabelling, sensor noise), not
adversarial model poisoning, where malicious clients craft their updates to steer the
global model. Class-consistent flips could change the observed relationship by concentrating corrupted
labels on one alternative class, unlike the scattered random flips tested here. The generalisability of the observed behaviour should therefore be tested across
additional architectures, datasets, client counts and behaviours (including collusion),
local objectives (including plain cross-entropy), aggregation strategies, and corruption models.

Our protocol isolates the diagnostic signals rather than using them for
sample selection or aggregation. Acting on the prediction-label
mismatch is closely related to established label-noise sample selection, including
Confident Learning~\cite{confident2021} and the federated methods surveyed by
FNBench~\cite{fnbench2025}. Evaluating how these signals perform once embedded in
complete defence pipelines is a distinct question beyond the scope of this diagnostic
study. In practice, the corruption type may not be known a priori; our results therefore
motivate monitoring both entropy-based uncertainty and prediction-label loss rather than
selecting either signal alone. Designing and validating a combined decision rule is an
important follow-on problem.

\section{Conclusion}

This study asked which of two signals, input-conditional uncertainty and the
prediction-label loss, detects which kind of data corruption in FL, and why.
The two corruption types leave different signatures. Under the persistent random flips studied here, wrong labels are scattered across classes and the global model generally remains confident in the underlying image; consequently, input-conditional uncertainty does not systematically distinguish flipped from clean samples, while the prediction-label mismatch remains informative because it reads the supplied label. The within-client results confirm this separation: uncertainty remains at
chance on flipped samples (AUC $0.49$--$0.50$), whereas loss separates them
strongly ($0.85$ on CIFAR-10 and $0.95$ on SVHN). For image noise,
expected-entropy uncertainty is instead modestly more informative than loss, and its relative advantage becomes apparent as federation-wide corruption prevalence increases. This difference is important. Label corruption, which uncertainty fails to detect, reduces clean-test accuracy by up to 39 points, whereas Gaussian image noise, which expected entropy detects, has little adverse effect. These results caution against treating uncertainty as a universal FL data-quality signal and support prediction-label mismatch as the
primary diagnostic for the persistent random label corruption studied here.
Important next steps are testing whether corruption detectability depends on task difficulty using more challenging datasets, characterising when the separation between uncertainty and loss erodes as corrupted samples come to dominate a class, using these signals to build a federated corruption detector, and integrating them into an uncertainty-aware approach~\cite{Ho:FLTA2026}, where sample-level corruption information can inform aggregation weights.

\section*{Acknowledgement}
This work was supported by the UK Research and Innovation DARE UK Real-world Research Exemplar Programme (Ref:~UKRI4079).

\clearpage
\appendices
\section{Full Detection Grids}

These tables report detection AUC across the full corruption sweep and
per-client robustness checks.

\begin{table}[h!]
\centering
\caption{Within-client per-sample AUC by corruption type, $M$, and $q$;
each cell averages $\alpha\times$seed runs. Abbreviations as in
Table~\ref{tab:persample}.}
\label{tab:persample-full}
\footnotesize
\setlength{\tabcolsep}{1.5pt}
\renewcommand{\arraystretch}{0.78}
\begin{tabular*}{\columnwidth}{@{\extracolsep{\fill}}llc ccc ccc}
\toprule
 & & & \multicolumn{3}{c}{CIFAR-10} & \multicolumn{3}{c}{SVHN} \\
\cmidrule(lr){4-6}\cmidrule(lr){7-9}
Corrupt & $M$ & Signal & $q{=}0.2$ & $q{=}0.4$ & $q{=}0.6$
& $q{=}0.2$ & $q{=}0.4$ & $q{=}0.6$ \\
\midrule
\multirow{20}{*}{Flip}
 & \multirow{5}{*}{2} & AV & 0.47 & 0.48 & 0.47 & 0.50 & 0.48 & 0.50 \\
 & & AE & 0.48 & 0.48 & 0.49 & 0.53 & 0.49 & 0.47 \\
 & & EV & 0.52 & 0.48 & 0.48 & 0.52 & 0.49 & 0.46 \\
 & & EE & 0.52 & 0.48 & 0.48 & 0.52 & 0.49 & 0.46 \\
 & & L  & \textbf{0.93} & \textbf{0.90} & \textbf{0.89}
        & \textbf{0.98} & \textbf{0.98} & \textbf{0.97} \\
 & \multirow{5}{*}{4} & AV & 0.49 & 0.48 & 0.49 & 0.50 & 0.48 & 0.49 \\
 & & AE & 0.51 & 0.50 & 0.49 & 0.50 & 0.49 & 0.48 \\
 & & EV & 0.51 & 0.50 & 0.49 & 0.50 & 0.49 & 0.48 \\
 & & EE & 0.51 & 0.51 & 0.49 & 0.50 & 0.49 & 0.49 \\
 & & L  & \textbf{0.91} & \textbf{0.87} & \textbf{0.85}
        & \textbf{0.98} & \textbf{0.97} & \textbf{0.97} \\
 & \multirow{5}{*}{6} & AV & 0.49 & 0.49 & 0.49 & 0.49 & 0.49 & 0.49 \\
 & & AE & 0.50 & 0.49 & 0.49 & 0.50 & 0.50 & 0.50 \\
 & & EV & 0.50 & 0.50 & 0.50 & 0.49 & 0.50 & 0.50 \\
 & & EE & 0.50 & 0.50 & 0.50 & 0.49 & 0.50 & 0.50 \\
 & & L  & \textbf{0.87} & \textbf{0.83} & \textbf{0.79}
        & \textbf{0.97} & \textbf{0.95} & \textbf{0.93} \\
 & \multirow{5}{*}{8} & AV & 0.49 & 0.49 & 0.49 & 0.51 & 0.49 & 0.50 \\
 & & AE & 0.50 & 0.49 & 0.51 & 0.51 & 0.51 & 0.50 \\
 & & EV & 0.50 & 0.50 & 0.51 & 0.51 & 0.50 & 0.50 \\
 & & EE & 0.50 & 0.50 & 0.51 & 0.51 & 0.50 & 0.50 \\
 & & L  & \textbf{0.85} & \textbf{0.77} & \textbf{0.70}
        & \textbf{0.96} & \textbf{0.92} & \textbf{0.86} \\
\midrule
\multirow{20}{*}{Noise}
 & \multirow{5}{*}{2} & AV & 0.50 & 0.48 & 0.46 & 0.22 & 0.22 & 0.25 \\
 & & AE & 0.68 & 0.68 & 0.68 & \textbf{0.69} & \textbf{0.68} & \textbf{0.65} \\
 & & EV & 0.63 & 0.62 & 0.64 & 0.65 & 0.63 & 0.61 \\
 & & EE & 0.65 & 0.64 & 0.65 & 0.65 & 0.64 & 0.61 \\
 & & L  & \textbf{0.70} & \textbf{0.70} & \textbf{0.69}
        & 0.68 & 0.67 & \textbf{0.65} \\
 & \multirow{5}{*}{4} & AV & 0.56 & 0.50 & 0.49 & 0.24 & 0.27 & 0.29 \\
 & & AE & \textbf{0.73} & \textbf{0.69} & \textbf{0.66}
        & \textbf{0.69} & \textbf{0.66} & \textbf{0.64} \\
 & & EV & 0.62 & 0.60 & 0.61 & 0.65 & 0.62 & 0.60 \\
 & & EE & 0.64 & 0.61 & 0.62 & 0.65 & 0.62 & 0.60 \\
 & & L  & 0.69 & 0.66 & 0.63 & 0.67 & 0.65 & 0.62 \\
 & \multirow{5}{*}{6} & AV & 0.60 & 0.56 & 0.52 & 0.29 & 0.29 & 0.32 \\
 & & AE & \textbf{0.73} & \textbf{0.67} & \textbf{0.63}
        & \textbf{0.67} & \textbf{0.64} & \textbf{0.62} \\
 & & EV & 0.62 & 0.58 & 0.57 & 0.63 & 0.61 & 0.59 \\
 & & EE & 0.64 & 0.59 & 0.58 & 0.63 & 0.61 & 0.59 \\
 & & L  & 0.66 & 0.63 & 0.60 & 0.65 & 0.63 & 0.61 \\
 & \multirow{5}{*}{8} & AV & 0.59 & 0.54 & 0.50 & 0.31 & 0.33 & 0.34 \\
 & & AE & \textbf{0.70} & \textbf{0.63} & \textbf{0.60}
        & \textbf{0.66} & \textbf{0.64} & \textbf{0.61} \\
 & & EV & 0.59 & 0.56 & 0.54 & 0.62 & 0.60 & 0.58 \\
 & & EE & 0.60 & 0.57 & 0.55 & 0.62 & 0.60 & 0.58 \\
 & & L  & 0.64 & 0.60 & 0.57 & 0.64 & 0.63 & 0.60 \\
\bottomrule
\end{tabular*}
\renewcommand{\arraystretch}{1.0}
\end{table}

\begin{table}[h!]
\centering
\caption{Per-client AUC by Dirichlet $\alpha$, averaged over the
$M\times q\times$seed grid. Abbreviations as in Table~\ref{tab:perclient}.}
\label{tab:perclient-alpha}
\footnotesize
\setlength{\tabcolsep}{1.2pt}
\renewcommand{\arraystretch}{0.70}
\begin{tabular*}{\columnwidth}{@{\extracolsep{\fill}}llcccccc}
\toprule
Dataset & Corrupt & $\alpha$ & AV & AE & EV & EE & $1{-}$Acc \\
\midrule
CIFAR-10 & Flip  & 0.1 & 0.37 & 0.69 & 0.55 & 0.60 & \textbf{0.91} \\
         &       & 0.3 & 0.34 & 0.60 & 0.52 & 0.55 & \textbf{0.98} \\
         &       & 1.0 & 0.53 & 0.61 & 0.57 & 0.61 & \textbf{1.00} \\
CIFAR-10 & Noise & 0.1 & 0.45 & \textbf{0.76} & 0.65 & 0.71 & 0.67 \\
         &       & 0.3 & 0.33 & \textbf{0.87} & 0.66 & 0.71 & 0.76 \\
         &       & 1.0 & 0.59 & \textbf{0.92} & 0.78 & 0.84 & 0.77 \\
\midrule
SVHN     & Flip  & 0.1 & 0.45 & 0.68 & 0.60 & 0.62 & \textbf{0.97} \\
         &       & 0.3 & 0.41 & 0.69 & 0.58 & 0.59 & \textbf{1.00} \\
         &       & 1.0 & 0.50 & 0.82 & 0.65 & 0.67 & \textbf{1.00} \\
SVHN     & Noise & 0.1 & 0.33 & 0.72 & 0.69 & 0.69 & \textbf{0.75} \\
         &       & 0.3 & 0.26 & \textbf{0.98} & 0.84 & 0.85 & 0.96 \\
         &       & 1.0 & 0.08 & \textbf{1.00} & 0.94 & 0.95 & 0.99 \\
\bottomrule
\end{tabular*}
\renewcommand{\arraystretch}{1.0}
\end{table}

\begin{table}[t]
\centering
\caption{Per-client AUC by corruption type, $M$, and $q$; each cell
averages $\alpha\times$seed runs. Abbreviations as in
Table~\ref{tab:perclient}.}
\label{tab:perclient-full}
\footnotesize
\setlength{\tabcolsep}{1.5pt}
\renewcommand{\arraystretch}{0.76}
\begin{tabular*}{\columnwidth}{@{\extracolsep{\fill}}llc ccc ccc}
\toprule
 & & & \multicolumn{3}{c}{CIFAR-10} & \multicolumn{3}{c}{SVHN} \\
\cmidrule(lr){4-6}\cmidrule(lr){7-9}
Corrupt & $M$ & Signal & $q{=}0.2$ & $q{=}0.4$ & $q{=}0.6$
& $q{=}0.2$ & $q{=}0.4$ & $q{=}0.6$ \\
\midrule
\multirow{20}{*}{Flip}
 & \multirow{5}{*}{2} & AV & 0.46 & 0.41 & 0.41 & 0.52 & 0.57 & 0.54 \\
 & & AE & 0.59 & 0.58 & 0.61 & 0.75 & 0.74 & 0.73 \\
 & & EV & 0.53 & 0.53 & 0.58 & 0.66 & 0.67 & 0.67 \\
 & & EE & 0.54 & 0.58 & 0.61 & 0.67 & 0.68 & 0.68 \\
 & & $1{-}$Acc & \textbf{0.92} & \textbf{0.98} & \textbf{1.00}
              & \textbf{0.99} & \textbf{1.00} & \textbf{1.00} \\
 & \multirow{5}{*}{4} & AV & 0.34 & 0.37 & 0.33 & 0.48 & 0.53 & 0.56 \\
 & & AE & 0.59 & 0.64 & 0.68 & 0.72 & 0.72 & 0.72 \\
 & & EV & 0.48 & 0.53 & 0.56 & 0.59 & 0.60 & 0.65 \\
 & & EE & 0.53 & 0.57 & 0.64 & 0.59 & 0.62 & 0.67 \\
 & & $1{-}$Acc & \textbf{0.92} & \textbf{0.97} & \textbf{0.99}
              & \textbf{0.98} & \textbf{1.00} & \textbf{1.00} \\
 & \multirow{5}{*}{6} & AV & 0.49 & 0.42 & 0.48 & 0.40 & 0.42 & 0.50 \\
 & & AE & 0.68 & 0.66 & 0.63 & 0.69 & 0.74 & 0.73 \\
 & & EV & 0.49 & 0.56 & 0.59 & 0.55 & 0.63 & 0.66 \\
 & & EE & 0.55 & 0.59 & 0.58 & 0.58 & 0.65 & 0.68 \\
 & & $1{-}$Acc & \textbf{0.91} & \textbf{0.98} & \textbf{0.99}
              & \textbf{0.97} & \textbf{1.00} & \textbf{1.00} \\
 & \multirow{5}{*}{8} & AV & 0.41 & 0.43 & 0.40 & 0.27 & 0.32 & 0.32 \\
 & & AE & 0.64 & 0.68 & 0.63 & 0.71 & 0.77 & 0.74 \\
 & & EV & 0.53 & 0.61 & 0.57 & 0.54 & 0.59 & 0.55 \\
 & & EE & 0.57 & 0.65 & 0.61 & 0.53 & 0.60 & 0.58 \\
 & & $1{-}$Acc & \textbf{0.90} & \textbf{0.98} & \textbf{1.00}
              & \textbf{0.96} & \textbf{0.99} & \textbf{1.00} \\
\midrule
\multirow{20}{*}{Noise}
 & \multirow{5}{*}{2} & AV & 0.49 & 0.45 & 0.37 & 0.31 & 0.20 & 0.13 \\
 & & AE & \textbf{0.79} & \textbf{0.90} & 0.95
        & 0.94 & 0.98 & \textbf{0.98} \\
 & & EV & 0.75 & 0.82 & 0.92 & 0.88 & 0.96 & \textbf{0.98} \\
 & & EE & \textbf{0.79} & 0.85 & \textbf{0.96}
        & 0.87 & 0.97 & \textbf{0.98} \\
 & & $1{-}$Acc & 0.73 & 0.84 & 0.92 & \textbf{0.97}
              & \textbf{1.00} & 0.97 \\
 & \multirow{5}{*}{4} & AV & 0.42 & 0.40 & 0.33 & 0.39 & 0.25 & 0.16 \\
 & & AE & \textbf{0.85} & \textbf{0.93} & \textbf{0.95}
        & \textbf{0.88} & \textbf{0.93} & 0.95 \\
 & & EV & 0.66 & 0.69 & 0.79 & 0.82 & 0.88 & 0.90 \\
 & & EE & 0.72 & 0.75 & 0.85 & 0.82 & 0.88 & 0.90 \\
 & & $1{-}$Acc & 0.68 & 0.78 & 0.85 & 0.87
              & \textbf{0.93} & \textbf{0.97} \\
 & \multirow{5}{*}{6} & AV & 0.56 & 0.59 & 0.55 & 0.35 & 0.22 & 0.16 \\
 & & AE & \textbf{0.83} & \textbf{0.89} & \textbf{0.90}
        & \textbf{0.86} & 0.89 & 0.91 \\
 & & EV & 0.63 & 0.67 & 0.76 & 0.72 & 0.81 & 0.87 \\
 & & EE & 0.70 & 0.77 & 0.81 & 0.75 & 0.82 & 0.86 \\
 & & $1{-}$Acc & 0.67 & 0.73 & 0.77 & 0.83
              & \textbf{0.90} & \textbf{0.92} \\
 & \multirow{5}{*}{8} & AV & 0.48 & 0.43 & 0.38 & 0.26 & 0.15 & 0.11 \\
 & & AE & \textbf{0.73} & \textbf{0.73} & \textbf{0.76}
        & \textbf{0.79} & \textbf{0.83} & 0.84 \\
 & & EV & 0.52 & 0.55 & 0.63 & 0.60 & 0.66 & 0.74 \\
 & & EE & 0.56 & 0.61 & 0.66 & 0.65 & 0.72 & 0.74 \\
 & & $1{-}$Acc & 0.61 & 0.61 & 0.64 & 0.77 & 0.82 & \textbf{0.85} \\
\bottomrule
\end{tabular*}
\renewcommand{\arraystretch}{1.0}
\end{table}

\begin{table}[h!]
\centering
\caption{Seed-specific grid averages of within-client image-noise AUC;
each row averages the 36 $M\times q\times\alpha$ configurations.
Differences are computed before rounding.}
\label{tab:noise-seed-values}
\footnotesize
\setlength{\tabcolsep}{1.5pt}
\renewcommand{\arraystretch}{0.78}
\begin{tabular*}{\columnwidth}{@{\extracolsep{\fill}}lcccc}
\toprule
Dataset & Seed & AE & Loss & AE$-$Loss \\
\midrule
CIFAR-10 & 42  & \textbf{0.697} & 0.656 & +0.041 \\
CIFAR-10 & 123 & \textbf{0.650} & 0.633 & +0.016 \\
CIFAR-10 & 777 & \textbf{0.675} & 0.646 & +0.029 \\
\midrule
SVHN & 42  & \textbf{0.662} & 0.648 & +0.014 \\
SVHN & 123 & \textbf{0.647} & 0.636 & +0.011 \\
SVHN & 777 & \textbf{0.659} & 0.643 & +0.016 \\
\bottomrule
\end{tabular*}
\renewcommand{\arraystretch}{1.0}
\end{table}


\begin{thebibliography}{00}
\bibitem{mcmahan2017} B. McMahan, E. Moore, D. Ramage, S. Hampson, and
B. Ag\"uera y Arcas, ``Communication-efficient learning of deep networks from
decentralized data,'' in \emph{Proc. AISTATS}, 2017, pp. 1273--1282.
\bibitem{fedned2024} Y. Lu et al., ``Federated learning with extremely noisy clients via negative distillation,'' in \emph{Proc. AAAI}, vol. 38, no. 13, 2024, pp. 14184--14192.
\bibitem{feddpso2025} C. Ouyang et al., ``Federated learning for extreme label noise: Enhanced knowledge distillation and particle swarm optimization,'' \emph{Electronics}, vol. 14, no. 2, Art. no. 366, 2025.
\bibitem{surefed2023} N. Heydaribeni, R. Zhang, T. Javidi, C. Nita-Rotaru, and
F. Koushanfar, ``SureFED: Robust federated learning via uncertainty-aware inward
and outward inspection,'' arXiv:2308.02747, 2023.
\bibitem{kendall2017} A. Kendall and Y. Gal, ``What uncertainties do we need in
Bayesian deep learning for computer vision?,'' in \emph{Proc. NeurIPS}, 2017,
pp. 5574--5584.
\bibitem{depeweg2018} S. Depeweg, J. M. Hern\'andez-Lobato, F. Doshi-Velez, and
S. Udluft, ``Decomposition of uncertainty in Bayesian deep learning for
efficient and risk-sensitive learning,'' in \emph{Proc. ICML}, 2018,
pp. 1184--1193.
\bibitem{gal2016} Y. Gal and Z. Ghahramani, ``Dropout as a Bayesian
approximation: Representing model uncertainty in deep learning,'' in
\emph{Proc. ICML}, 2016, pp. 1050--1059.
\bibitem{confident2021} C. G. Northcutt, L. Jiang, and I. L. Chuang, ``Confident
learning: Estimating uncertainty in dataset labels,'' \emph{J. Artif. Intell.
Res.}, vol. 70, pp. 1373--1411, 2021.
\bibitem{unicon2022} N. Karim, M. N. Rizve, N. Rahnavard, A. Mian, and M. Shah,
``UNICON: Combating label noise through uniform selection and contrastive
learning,'' in \emph{Proc. CVPR}, 2022, pp. 9676--9686.
\bibitem{uqled2024} J. Jakubik, M. V\"ossing, M. Maskey, C. W\"olfle, and
G. Satzger, ``Improving label error detection and elimination with uncertainty
quantification,'' \emph{J. Artif. Intell. Res.}, vol. 84, Art. no. 19, 2025.
\bibitem{ulc2022} Y. Huang, B. Bai, S. Zhao, K. Bai, and F. Wang,
``Uncertainty-aware learning against label noise on imbalanced datasets,'' in
\emph{Proc. AAAI}, vol. 36, no. 6, 2022, pp. 6960--6969.
\bibitem{fnbench2025} X. Jiang et al., ``FNBench: Benchmarking robust federated learning
against noisy labels,'' arXiv:2505.06684, 2025.
\bibitem{lfighter2024} N. M. Jebreel, J. Domingo-Ferrer, D. S\'anchez, and
A. Blanco-Justicia, ``LFighter: Defending against the label-flipping attack in
federated learning,'' \emph{Neural Networks}, vol. 170, pp. 111--126, 2024.
\bibitem{fedbary2024} W. Li, S. Fu, F. Zhang, and Y. Pang, ``Data valuation and
detections in federated learning,'' in \emph{Proc. CVPR}, 2024, pp. 12027--12036.
\bibitem{fedivon} S. Pal, A. Gupta, S. Sarwar, and P. Rai, ``Federated learning with
uncertainty and personalization via efficient second-order optimization,'' \emph{Trans.
Mach. Learn. Res.}, 2025 (arXiv:2411.18385).
\bibitem{fedpn2024} N. Kotelevskii, S. Horvath, K. Nandakumar, M. Takac, and
M. Panov, ``Dirichlet-based uncertainty quantification for personalized
federated learning with improved posterior networks,'' in \emph{Proc. IJCAI},
2024, pp. 7127--7135.
\bibitem{fedee} Y. Zhang, T. Xia, A. Ghosh, and C. Mascolo, ``Uncertainty-aware
personalized federated learning for realistic healthcare applications,'' in \emph{Proc.
Machine Learning for Health (ML4H)}, PMLR vol. 259, 2025, pp. 1067--1086.
\bibitem{udjfl2025} A. N. Carey and X. Wu, ``Achieving distributive justice in
federated learning via uncertainty quantification,'' \emph{IEEE Trans. Big Data},
vol. 12, no. 3, pp. 1058--1069, 2026.
\bibitem{udfgma2025} Z. Luo, A. Gooya, and E. S. L. Ho, ``UDF-GMA: Uncertainty
disentanglement and fusion for general movement assessment,'' \emph{IEEE J. Biomed. Health
Inform.}, vol.~30, no.~1, pp.~366--375, 2026.
\bibitem{bald2011} N. Houlsby, F. Husz\'ar, Z. Ghahramani, and M. Lengyel,
``Bayesian active learning for classification and preference learning,''
arXiv:1112.5745, 2011.
\bibitem{arpit2017} D. Arpit et al., ``A closer look at memorization in deep
networks,'' in \emph{Proc. 34th Int. Conf. Mach. Learn. (ICML)}, PMLR,
vol. 70, 2017, pp. 233--242.
\bibitem{he2016} K. He, X. Zhang, S. Ren, and J. Sun,
``Deep residual learning for image recognition,'' in \emph{Proc. CVPR},
2016, pp. 770--778.
\bibitem{krizhevsky2009} A. Krizhevsky, ``Learning multiple layers of features
from tiny images,'' Univ. of Toronto, Tech. Rep., 2009.
\bibitem{netzer2011} Y. Netzer, T. Wang, A. Coates, A. Bissacco, B. Wu,
and A. Y. Ng, ``Reading digits in natural images with unsupervised feature
learning,'' in \emph{NIPS Workshop Deep Learning and Unsupervised Feature
Learning}, 2011.

\bibitem{beutel2020flower} D. J. Beutel et al., ``Flower: A friendly federated
learning research framework,'' arXiv:2007.14390, 2020.



\bibitem{Ho:FLTA2026} E. S. L. Ho, ``Uncertainty-aware federated learning for infant movement analysis,'' in \emph{Proc. 4th Int. Conf. Federated Learning Technologies and Applications (FLTA)}, 2026, in press.

\end{thebibliography}
\end{document}